\documentclass[12pt]{article} 

\usepackage{amsfonts,amscd,amssymb}
\usepackage{amsthm,amsmath}
\usepackage{algorithm}
\usepackage[noend]{algpseudocode}

\usepackage[margin=1in]{geometry}

\usepackage{bm}
\usepackage{bbm} 
\usepackage{color}
\usepackage{verbatim}
\usepackage{graphicx}
\usepackage{setspace}
\usepackage{enumitem}
\usepackage{lipsum}

\usepackage{mathtools}
\usepackage{fixmath}
\usepackage{hyperref}

\usepackage{tikz}
\usetikzlibrary{matrix,calc}

\newtheorem{thm}{Theorem}
\newtheorem{lemma}{Lemma}
\newtheorem{prop}{Proposition}
\newtheorem{cor}{Corollary}
\newtheorem{remark}{Remark}

\newtheorem{mydef}{Definition}

\newcommand{\mb}{\mathbf}
\newcommand{\mbb}{\mathbb}
\newcommand{\mc}{\mathcal}

\newcommand{\argmin}{\operatornamewithlimits{argmin}}

\def\d{\textnormal{d}}
\def\dt{\textnormal{dt}}
\def\R{\mathbb{R}}
\def\x{\mathbf{x}}
\def\b{\mathbf{b}}

\def\w{\mathbf{W}}

\renewcommand{\mathbf}[1]{\mathbold{#1}}           
\newcommand{\mat}[1]{\mathbold{#1}}             
\DeclarePairedDelimiterX{\inp}[2]{\langle}{\rangle}{#1, #2} 
\newcommand{\T}{\top}                           
\newcommand{\ftimes}{\star}                     

\newcommand{\grad}{\nabla}                      
\newcommand{\partder}[2]{\frac{\partial #1}{\partial #2}} 

\newcommand{\E}[2][]{\mathbb{E}_{#1}\left[#2\right]} 
\DeclarePairedDelimiter\norm{\lVert}{\rVert}    
\DeclareMathOperator{\prox}{prox}               
\DeclareMathOperator{\diag}{diag}               
\DeclareMathOperator{\dist}{dist}               

\renewcommand{\L}{\mathcal{L}}                  
\renewcommand{\l}{\ell}                         
\newcommand{\half}{\frac{1}{2}}                 

\usepackage{mathtools}

\makeatletter
\DeclareRobustCommand\widecheck[1]{{\mathpalette\@widecheck{#1}}}
\def\@widecheck#1#2{%
    \setbox\z@\hbox{\m@th$#1#2$}%
    \setbox\tw@\hbox{\m@th$#1%
       \widehat{%
          \vrule\@width\z@\@height\ht\z@
          \vrule\@height\z@\@width\wd\z@}$}%
    \dp\tw@-\ht\z@
    \@tempdima\ht\z@ \advance\@tempdima2\ht\tw@ \divide\@tempdima\thr@@
    \setbox\tw@\hbox{%
       \raise\@tempdima\hbox{\scalebox{1}[-1]{\lower\@tempdima\box
\tw@}}}%
    {\ooalign{\box\tw@ \cr \box\z@}}}
\makeatother
\title{Adaptive Group Lasso Neural Network Models for Functions of Few Variables and Time-Dependent Data}
\date{\today}
\author{
Lam Si Tung Ho \\
Department of Mathematics and Statistics \\
Dalhousie University, Halifax, Nova Scotia, Canada
\and
Nicholas Richardson\\
Department of Applied Mathematics, University of Waterloo
\and
Giang Tran \\
Department of Applied Mathematics, University of Waterloo
}

\begin{document}
\maketitle

\begin{abstract}
In this paper, we propose an adaptive group Lasso deep neural network for high-dimensional function approximation where input data are generated from a dynamical system and the target function depends on few active variables or few linear combinations of variables.
We approximate the target function by a deep neural network and enforce an adaptive group Lasso constraint to the weights of a suitable hidden layer in order to represent the constraint on the target function.
We utilize the proximal algorithm to optimize the penalized loss function.
Using the non-negative property of the Bregman distance, we prove that the proposed optimization procedure achieves loss decay.
Our empirical studies show that the proposed method outperforms recent state-of-the-art methods including the sparse dictionary matrix method, neural networks with or without group Lasso penalty.
\end{abstract}

 \clearpage
\section{Introduction}

Function approximation is a fundamental problem that lies at the interface between applied mathematics and machine learning. The main goal of this problem is to approximate an unknown underlying function from a collection of inputs and outputs. Typically, we assume the unknown function belongs to a certain function space such as the spaces of polynomials, trigonometric functions, or Fourier terms. The underlying function can be written as a linear combination of those basis terms and the unknown coefficients can be found via solving a least square fitting problem. In another direction, we can approximate the underlying function as a composition of linear and simple nonlinear functions, namely neural networks \cite{cybenko1989approximation, barron1992neural}. Recently, deep neural networks have been shown as one of the best methods for function approximation \cite{cohen2016expressive, eldan2016power, liang2016deep, poggio2017, telgarsky2016benefits}.

Approximating high-dimensional functions is more challenging. The general idea for handling high-dimensional data is to add a regularization term such as the Tikhonov, $\ell_1$, or $\ell_0$-like penalties in order to capture active variables. For example, polynomial approximations with sparse coefficients have been proven theoretically and numerically to be one of the most effective dictionary methods in the high-dimensional setting \cite{rauhut2012sparse, brunton2016discovering, adcock2017compressed,chkifa2018polynomial, schaeffer2018extracting, ho2020recovery}. Neural networks, on the other hand, often utilize group penalties to force all weights associated with an inactive variable to zero together. For example, in \cite{scardapane2017group}, the authors propose to use group penalties for the weights in all layers of the neural network model to identify important features and prune the network simultaneously. 
Smoothing group $\ell_1$-penalties have been applied to the weights of the hidden layer in a single-hidden-layer neural network to identify useful features \cite{zhang2019feature}. 
In \cite{NEURIPS2020_1959eb9d}, the authors propose to use adaptive group Lasso for analytic deep neural networks.
Their simulations suggest that adaptive group Lasso is better than group Lasso in removing inactive variables.

Another important aspect of the function approximation problem is the properties of data. The most common setting assumes that data are independent and identically distributed (iid). However, the independence assumption does not hold for many applications such as epidemiology, evolutionary biology, financial prediction, and signal processing.
Therefore, much effort has been done to study function approximation for non-iid data. One popular type of non-iid data that has been extensively studied is mixing sequences \cite{steinwart2009learning, cuong2013generalization, hang2014fast, dinh2015learning, ho2020recovery, wong2020lasso}. However, mixing property is not a natural assumption to make and verifying this property is often difficult. Time-dependent data is another popular type of non-iid data, such as data generated from a dynamical system. There are connections between dynamical systems and mixing properties but figuring them out is extremely difficult and worth a separate study. There is a great interest recently in learning the underlying governing equations from time-dependent data using both dictionary methods \cite{brunton2016discovering, tran2017exact, raissi2018hidden, lu2019nonparametric} and the neural network methods \cite{qin2019data, sun2020neupde}. 

In this paper, we focus on the function approximation problem when data are generated from a high-dimensional dynamical system and the unknown smooth nonlinear function depends on few active variables. We propose to approximate the target function by a deep neural network with an adaptive group Lasso constraint to the weights of the first hidden layer. 
It has been proven recently that adaptive group Lasso can correctly identify active variables with high probability \cite{NEURIPS2020_1959eb9d}.
We also extend our method to the case when the target function depends on few linear combinations of variables, which is related to active subspace methods. 
For optimizing the regularized loss function, we apply the popular proximal algorithm \cite{parikh2014proximal}.
Using the non-negative property of the Bregman distance, we show that our proposed optimization procedure achieves loss decay.
Specifically, if the learning rate is sufficiently small, the regularized loss always improves after each iteration and the estimated value of the parameters converges to the set of critical points of the regularized loss function.
We conduct an extensive empirical study to compare the performance of the proposed method with the neural network models (with/without group Lasso penalty) and dictionary matrix method with the $\ell_1$-constraint to the coefficients. Our experiments demonstrate that neural networks with adaptive group Lasso outperform the other methods in learning various nonlinear functions where the input data are generated from a popular dynamical system, the Lorenz 96 system.

The paper is organized as follows. We explain the problem setting and present the proposed algorithm for learning functions of few active variables in Section \ref{sec:setting}.
We also prove the loss decay property of the proposed optimization algorithm here.
In Section \ref{sec:extension}, we show how to extend our method to the case when the target function depends on few linear combinations of variables and state the corresponding algorithm. Numerical results and comparisons are described in Section \ref{sec:numres}.

\section{Problem Setting and Algorithm}
\label{sec:setting}
\subsection{Problem Setting}
Consider the problem of learning an unknown nonlinear smooth function $f:\R^d\rightarrow \R$ from a set of input-output samples $\{(\x_i,y_i)\}_{i=1}^m$, where the input data $\{\x_i\}_{i=1}^m$ are (possibly noisy) time-dependent and the function $f$ depends on a few active variables, $f(\x) = f(\x\mid_{\mathcal{S}})$ with $\mathcal{S}\subset \{1,2,\ldots, d\}$. In our settings, we assume the input data are noisy observations of a high-dimensional dynamical system, \[\x_i = \widetilde{\x}(t_i) + \varepsilon_{\x}(t_i),\quad t_1<t_2<\cdots<t_m,\] 
where $\{\widetilde{\x}(t_i)\}_{i=1}^m$ are noiseless solutions of the dynamical system and the error $\varepsilon_\x$ is governed by the noise level $\sigma_\x$. The output data is also noisy with the error is governed by the noise level $\sigma_y$:
\[y_i = f(\widetilde{\x}(t_i)) +\varepsilon_y(t_i). \]
Our goal is to find a good approximation of the function $f$ and to accurately identify the active variables. 

\begin{figure}[htbp!]
\centering
\includegraphics[width = 6 in]{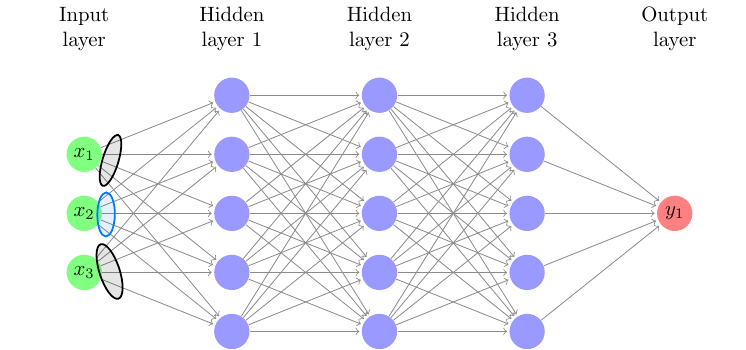}
\caption{An illustration of an adaptive group Lasso neural network: all weights connecting each input variable with nodes in the first hidden layer are grouped together (weights in the same circle are in the same group). The color of circles represent whether the associated variable is active (blue) or inactive (black).}
\label{fig:NN-architect}
\end{figure}

To achieve this goal, we first approximate the target function $f$ by a neural network: 
\[
f\approx F_{\theta }(\mb{x}) = \w_{L}(\sigma (\w_{L-1}(\cdots \sigma(\w_2( \sigma (\w_{1}\mb{x}+\b_1)) +\b_2)\cdots )+\b_{L-1}))+\b_L,
\]
where $L$ is the number of layers of the neural network, $\sigma$ is the $\tanh$, $\theta = \{(\w_l,\b_l)\}_{l=1}^L$ are trainable weights and biases. 
Moreover, since $f$ depends only on few active variables, we would like to enforce the weights connecting nonactive variables with the first hidden-layer to zero. One of the effective methods for this task is to use an adaptive sparse group Lasso penalty, where all weights in the first hidden layer associated with each input variable are grouped together. 
An illustration of an adaptive group Lasso neural network is presented in Figure \ref{fig:NN-architect}.
In \cite{NEURIPS2020_1959eb9d}, the authors consider a simpler scenario where $f$ is an analytic deep neural network and the approximated function $F_{\theta }(\mb{x})$ has the same structure as $f$.
Moreover, the inputs are iid and have no noise.
Under this setting, they prove that the adaptive sparse group Lasso can select the correct set of active variables with high probability.
As a consequence, we can recover the underlying function $f$ as the number of inputs increases.
Although the result is only for a simple case, it suggests that the adaptive sparse group Lasso may also work for our setting.
We will confirm this observation using simulation studies in Section \ref{sec:numres}.

Specifically, the adaptive sparse group Lasso aims to solve the following optimization problem:
\begin{equation}
\widehat{\theta} \in \argmin_{\theta} \mathcal{L}(\theta) + \mc{R}(\w_1),
\label{eqn:lasso}
\end{equation}
where
\[
 \mathcal{L}(\theta)= \dfrac{1}{m}\sum\limits_{i=1}^m |y_i - F_{\theta}(\x_i)|^2,\]
 and 
 \[\mc{R}(\w_1) = \lambda_m \sum_{j=1}^d{ \frac{1}{\| \widetilde{\w}_1{[:,j]}\|^2} \| \w_1{[:,j]} \|}.
\]
Here, $\lambda_m$ is a regularizer factor and $\w_1{[:,j]}$ denotes the $j$-th column of $\w_1$. Throughout this work, we use the Euclidean norm when we write $\norm{\cdot}$ without a trailing subscript. The initial estimator $\widetilde{\w}_1$ is the weight matrix of the first hidden layer of the neural network, which is obtained from solving the following non-linear least square problem:
\[
\widetilde{\theta}\in \argmin_{\theta} \mathcal{L}(\theta) .
\]
\subsection{Algorithm}
The optimization problem \eqref{eqn:lasso} can be solved using the proximal algorithm \cite{parikh2014proximal}:
\begin{equation}\label{eqn:proximal}
\theta^{(k + 1)} = \textbf{prox}_{\tau_k \mc{R}} \left[ \theta^{(k)} - \tau_k \nabla_{\theta} \mathcal{L}(\theta^{(k)} ) \right ]
\end{equation}
where
\begin{equation}
\textbf{prox}_{\tau_k \mc{R}}(\overline{\theta}) = \argmin_{\theta} \mc{R}(\w_1) + \frac{1}{2 \tau_k}\| \theta - \overline{\theta}\|_F^2.
\end{equation}
Denote $\overline{\theta} = \theta^{(k)} - \tau_k \nabla_{\theta} \mathcal{L}(\theta^{(k)} )$. Then the closed form solution of \eqref{eqn:proximal} is:
\[
\begin{aligned}
\w_1^{(k+1)}{[:,j]} &= \max \left (0, \| \overline{\w}_1{[:,j]} \| - \dfrac{\lambda_m \tau_k}{\|\widetilde{\w}_1{[:,j]}\|^2} \right )\frac{\overline{\w}_1{[:,j]}}{\| \overline{\w}_1{[:,j]} \|} ,\quad\text{for}\quad j = 1,\ldots, d,\\
\w_l^{(k+1)} &= \overline{\w}_l,\quad\text{for}\quad l =2, \ldots, L, \\
\b_l^{(k+1)} &= \overline{\b}_l,\quad \text{for}\quad l = 1,\ldots, L.
\end{aligned}
\]

Finally, in the preprocessing step, the training dataset is standardized by dividing each variable with the sample standard deviations of the corresponding variables. That is, 
\begin{equation}
y_i^{\text{new}}: = \frac{y_i}{\alpha}\quad\text{and}\quad x_{i,j}^{\text{new}} := \dfrac{x_{i,j}}{\sigma_j},\quad\text{for} \quad j=1,\ldots, d,
\end{equation}
where $\alpha$ is the sample standard deviation of the output dataset $\{y_i\}_{i=1}^m$, $\x_i=(x_{i,1},\ldots, x_{i,d})^T$, and $\sigma_j$ is the sample standard deviation of the $j$-th variable of the input dataset $\{x_{i,j}\}_{i=1}^m$. 
The method is summarized in Algorithm \ref{alg}.

\begin{algorithm}[htbp]
\caption{ }
\label{alg}
\begin{algorithmic}[1]
\State {\bfseries Input:} $m$ input-output pairs $\{(\x_i,y_i)\}_{i=1}^m\subset\R^d\times \R$.
\State{\bfseries Preprocess:} Standardize training data by their sample standard deviations:
\[y_i^{\text{new}}: = \frac{y_i}{\alpha}\quad\text{and}\quad x_{i,j}^{\text{new}} := \dfrac{x_{i,j}}{\sigma_j},\quad\text{for} \quad j=1,\ldots, d.\]

\Procedure{Find the initial estimators $\widetilde{\theta}$:}{}
\[\widetilde{\theta}\in\argmin_{\theta} \mathcal{L}(\theta) = \frac{1}{m}\sum_{i=1}^m{|y_i - F_{\theta}(\mb{x}_i)|^2}.\]
	\State{{\bf Initialize} $\widetilde{\theta}^{(0)}$} \For{$k<\text{iterMax}$}
	\State{$\widetilde{\theta}^{(k+1)} = Adam(\widetilde{\theta}^{(k)})$}
	\EndFor
\EndProcedure
\Procedure{Solve the Adaptive Group Lasso model:}{}
\[\widehat{\theta} \in \argmin_{\theta} \mathcal{L}(\theta) + \mathcal{R}(\w_1).
\]
	\State{{\bf Initialize} ${\theta}^{(0)}$}
	\While{$k< K$}
			\State {$\overline{\theta} =	\theta^{(k)} - \tau_k \nabla_{\theta} \mathcal{L}({\theta}^{(k)}  )$}
		\State{${\w}_1^{(k + 1)}[:,j]	= \max \left (0, \| \overline{\w}_1{[:,j]} \| - \dfrac{\lambda_m \tau_k}{\|\widetilde{\w}_1{[:,j]}\|^2} \right )\dfrac{\overline{\w}_1{[:,j]}}{\| \overline{\w}_1{[:,j]} \|},\quad\text{for}\ j=1,\ldots, d.$}
		\State ${\w}_l^{(k+1)}= \overline{\w}_l,\quad\text{for}\ l =2,\ldots, L.$
		\State ${\b}_l^{(k+1)} = \overline{\mb b}_l,\quad \text{for}\ l =1,\ldots, L.$
	\EndWhile{ }
\EndProcedure
\State {\bfseries Output:}  $\widehat{\theta} = \theta^{(K)}, \widehat{f}(\mb x) = F_{\widehat{\theta}}(\mb x)$; and the support set $\widehat{S} =\{ j: \| \widehat{\w}_1{[:,j]} \| > 0 \}.$
\end{algorithmic}
\end{algorithm}

\subsection{Loss Decay Property}
Denote $J(\theta) = \mathcal{R}(\w_1),\mathcal{F}(\theta)= \mathcal{L}(\theta) + J(\theta)$, and
\[ \mathbf{L}(\theta;B) =  \dfrac{1}{|B|}\sum\limits_{(\x_i,y_i)\in B} |y_i - F_{\theta}(\x_i)|^2,\]
where $B$ is a mini-batch of training data. In this section, we provide convergence analysis of the proximal gradient step in Algorithm~\ref{alg}. Specifically, we prove that under some mild assumptions, the total loss $\mathcal{F}(\theta)$ decays. We also illustrate the assumption verification for one-hidden layer networks.  

Using the non-negative property of the Bregman distance, we derive a slightly stronger result than the one from \cite{yun2020general}.
\begin{thm}\label{th:loss decay} Assume the following conditions:
\begin{enumerate}
    \item (Lipschitz Smoothness). The function $\nabla\mathcal{L}=\nabla_\theta\mathcal{L} (\theta) $ is Lipschitz with constant $C>0$:
    \[\|\nabla\mathcal{L} (\theta) - \nabla\mathcal{L}(\phi) \|\leq  C\|\theta - \phi\|.\]
    \item (Bounded Variance). There exists a constant $\mu>0$ such that for any trainable parameters $\theta$, we have
    \[\mathbb{E}\left[\|\nabla \mathbf{L}(\theta;B) - \nabla\mathcal{L}(\theta)\|^2\right] \leq \mu.\]
\end{enumerate}
Then the proximal stochastic gradient algorithm with the step sizes $\tau_k\leq \dfrac{1}{C}$ for all $k$,
\begin{align} \label{proxsgd}
\begin{split}
    g^{(k)}              &\leftarrow \grad_\theta{\mathbf{L}(\theta^{(k)};B^{(k)}}) \\
    \overline{\theta}^{(k+1)} &\leftarrow \theta^{(k)} - \tau_k g^{(k)}     \\
    \theta^{(k+1)}     &\leftarrow \prox_{\tau_k J} \left(\overline{\theta}^{(k+1)}\right),
\end{split}
\end{align}
achieves the total loss decay. That is,

 \begin{equation}\label{eqn:theorem1}
 \E{ \mathcal{F}(\theta^{(k+1)})}+   \left(\frac{1}{\tau_k} - C\right)\E{\norm{\theta^{(k+1)}-\theta^{(k)}}^2 }\leq \E{\mathcal{F}(\theta^{(k)})}+ \dfrac{\mu}{2C}.\end{equation}
In particular, if we use the full batch, then $\mu=0$ and we have the total loss decay:
\begin{equation} \mathcal{F}(\theta^{(k+1)}) +  \left(\dfrac{1}{\tau_k} - C\right) \norm{\theta^{(k+1)}-\theta^{(k)}}^2 \leq \mathcal{F}(\theta^{(k)}).\end{equation}
\end{thm}

\begin{proof} See Appendix \ref{sec:convergence proof0}. \end{proof}

Using Theorem \ref{th:loss decay}, we also achieve a similar estimation of the convergence of the stochastic proximal algorithm as in Theorem 1  \cite{yun2020general}.

\begin{thm} \label{thm:convergence}
Assume that $\tau_k\in [\gamma_1,\gamma_2]$, where $0<\gamma_1<\gamma_2<\dfrac{1}{C}$ and all assumptions in Theorem \ref{th:loss decay} are satisfied. Then
\begin{equation}
    \frac{1}{K}\sum_{k=0}^{K-1}\mathbb{E} \left[\dist(0, \partial \mathcal{F}(\theta^{(k+1)}))^2\right]
     \leq \left(3+\frac{3B}{2C}\right)\mu + \frac{3B}{K} \mathbb{E}[\mathcal{F}(\theta^{(0)}) - \mathcal{F}(\theta^{(K)})],
     \end{equation}
where  $B = \max
\limits_{t \in [\gamma_1,\gamma_2]}\left(\frac{t^{-2} + C^2}{t^{-1} - C}\right).$
\end{thm}

\begin{proof} See Appendix \ref{sec:convergence proof}.\end{proof}

Finally, we want to verify that in certain situations, the function $\mathcal{L}(\theta)$ satisfies the Lipschitz smooth assumption of Theorems (\ref{th:loss decay},\ref{thm:convergence}).
\begin{thm}\label{thm3}
Consider a one-hidden layer network, \[ F_\theta(\mathbf{x}) = \w_2\sigma(\w_1\mathbf{x}+\mathbf{b}_1) +\mathbf{b}_2,\]
where $\w_1\in \R^{h\times d}, \w_2\in \R^{n\times h} ,\mathbf{b}_1\in \R^h, \mathbf{b}_2\in \R^n, \mathbf{x}\in \R^d, \mathbf{y}\in \R^n$, and $\sigma$ is a one-Lipschitz smooth non-linear activation function. Then the associated mean squared error loss $\L(\theta)$ over all data points $\{(\mathbf{x}_i,\mathbf{y}_i)\}_{i=1}^m$ is $C$-smooth with
\begin{equation}
    C  = \max_{1\leq i\leq m} 2(h+1) \left( C_1\sqrt{n} \norm{\check{\mathbf{x}_i}}_2 + 1\right) \sqrt{C_1^2n(h+1) \|\check{\mathbf x}_i\|_2^2 +1}= \mathcal{O}(nC_1^2h^{3/2}\|\check{\mathbf x}_i\|_2^2)
\end{equation}
where $C_1$ is an upper bound for the magnitude of the entries in the outer layer and $\norm{\mathbf{\check{x}}}_2=\sqrt{\norm{\mathbf{x}}_2^2+1}$.
\end{thm}
\begin{proof} See Appendix \ref{sec:lipschitz proof}.\end{proof}

\begin{remark}
Hyperbolic tangent, $\tanh(x)$, and sigmoid, $(1+e^{-x})^{-1}$, are examples of 1-Lipschitz and 1-smooth non-linear activation functions since the magnitude of their first and second derivatives are bounded above by 1.
\end{remark}

\section{Extension to $f(\mb x) = g(A \mb x)$}
\label{sec:extension}

Next, we propose an adaptation of our proposed method to the scenario where the underlying function $f(\x)$ depends only on few linear combinations of variables.
Mathematically, the problem can be formulated as finding the function of the form $f(\mb x) = g(A \mb x)$ where $A$ is an unknown $k \times d$ matrix with $k \ll d$ and $g: \mbb{R}^k \to R$ is an unknown smooth function. We note that $k$ is also an unknown quantity.

Adapting the method in Section \ref{sec:setting}, we approximate $f$ by the following neural network
\[
f \approx G_{\theta}(\mb{x}) = \w_{L}(\sigma (\w_{L-1}(\cdots \sigma(\w_2( \sigma_1 (\w_{1}\mb{x}+\b_1)) +\b_2)\cdots )+\b_{L-1}))+\b_L,
\]
where $\sigma_1$ is the identity mapping and $\sigma$ is the $\tanh$.
Since $f$ depends only on few linear combinations of $\mb x$, we would like to enforce the weights connecting nonactive variables with the second hidden-layer to zero.\\
The corresponding optimization problem is
\begin{equation}
\widehat{\theta} \in \argmin_{\theta} \dfrac{1}{m}\sum\limits_{i=1}^m |y_i - G_{\theta}(\x_i)|^2 + \mc{R}(\w_2).
\label{eqn:lasso2}
\end{equation}
where
\[
\mc{R}(\w_2) = \lambda_m \sum_{j=1}^d{ \frac{1}{\| \widetilde{\w}_2{[:,j]}\|^2} \| \w_2{[:,j]} \|},
\]
and the initial estimator $\widetilde{\w}_2$ is obtained from solving the following non-linear least square problem:
\[
\widetilde{\theta}\in \argmin_{\theta}\left( \frac{1}{m}\sum_{i=1}^m{|y_i - G_{\theta}(\mb{x}_i)|^2}\right).
\]
Again, we can use the proximal algorithm (see Equation \eqref{eqn:proximal}) to solve the optimization problem \eqref{eqn:lasso2}. Specifically, the weights and biases in all layers except the second layer are updated by using a step of gradient descent. On the other hand, the weight matrix in the second layer is updated using the proximal solution:
\[
\begin{aligned}
&\overline{\w}_2 = \w_2^{(k)} -\tau_k \dfrac{\partial}{\partial \w_2}\mathcal{L}(\theta^{(k)}),\\
&{\w}_2^{(k + 1)}[:,j] = \max \left (0, \| \overline{\w}_2{[:,j]} \| - \dfrac{\lambda_m \tau_k}{\|\widetilde{\w}_2{[:,j]}\|^2} \right )\dfrac{\overline{\w}_2{[:,j]}}{\| \overline{\w}_2{[:,j]} \|}.
\end{aligned}
\]
We summarize the method in Algorithm \ref{alg2}.
\begin{algorithm}[htbp!]
\caption{ }
\label{alg2}
\begin{algorithmic}[1]
\State {\bfseries Input:} $m$ input-output pairs $\{(\x_i,y_i)\}_{i=1}^m\subset\R^d\times \R$.
\State{\bfseries Preprocess:} Standardize training data by their sample standard deviations:
\[y_i^{\text{new}}: = \frac{y_i}{\alpha}\quad\text{and}\quad x_{i,j}^{\text{new}} := \dfrac{x_{i,j}}{\sigma_j},\quad\text{for} \quad j=1,\ldots, d.\]

\Procedure{Find the initial estimators $\widetilde{\theta}$:}{}
\[\widetilde{\theta}\in\argmin_{\theta} \mathcal{L}(\theta) = \frac{1}{m}\sum_{i=1}^m{|y_i - G_{\theta}(\mb{x}_i)|^2}.\]
	\State{{\bf Initialize} $\widetilde{\theta}^{(0)}$} 
	\For{$k<\text{iterMax}$}
	\State{$\widetilde{\theta}^{(k+1)} = Adam(\widetilde{\theta}^{(k)})$}
	\EndFor
\EndProcedure
\Procedure{Solve the Adaptive Group Lasso model:}{}
\[\widehat{\theta} \in \argmin_{\theta} \mathcal{L}(\theta) + \mathcal{R}(\w_2) 
\]
	\State{{\bf Initialize} ${\theta}^{(0)}$} 
	\While{$k<\text{epochMax}$}
			\State {$\overline{\theta} =	\theta^{(k)} - \tau_k \nabla_{\theta} \mathcal{L}(\theta^{(k)})$}

		\State{${\w}_2^{(k + 1)}[:,j]	= \max \left (0, \| \overline{\w}_2{[:,j]} \| - \dfrac{\lambda_m \tau_k}{\|\widetilde{\w}_2{[:,j]}\|^2} \right )\dfrac{\overline{\w}_2{[:,j]}}{\| \overline{\w}_2{[:,j]} \|},\quad\text{for}\ j=1,\ldots,d. 
		$}
		\State${\w}_l^{(k+1)}= \overline{\w}_l,\quad\text{for}\ l=1,3,4\ldots,L.$
		\State$\b_l^{(k+1)} = \overline{\mb b}_l$,\quad \text{for}\quad l = 1,\ldots, L.
	\EndWhile{ }
\EndProcedure
\State {\bfseries Output:} $\widehat{\theta} =\theta^{(K)}$ and $\widehat{f}(\mb x) = G_{\widehat{\theta}}(\mb{x})$.
\end{algorithmic}
\end{algorithm}

\section{Numerical Results}
\label{sec:numres}
In this section, we demonstrate the performance of our proposed algorithm for input data generated from high-dimensional systems of ODEs. 
We apply our method with 3-hidden-layer neural networks where each hidden layer has $H$ nodes.
We measure the performance of our proposed method using sensitivity, specificity, and relative test error.
The sensitivity and specificity are defined as follows:
\begin{align*}
\text{sensitivity} &= \frac{\emph{true positive}}{\emph{positive}}, \\
\text{specificity} &= \frac{\emph{true negative}}{\emph{negative}},
\end{align*}
where \emph{true positive} is the number of variables that are correctly selected, \emph{true negative} is the number of variables that are correctly not selected, \emph{positive} is the number of variables in the true model, and \emph{negative} is the number of variables that are not in the true model.
For a test set $\mc{T}$, the relative test error of an algorithm is defined as follows:
\[
\sqrt{\frac{\sum_{\mb{x} \in \mc{T}}\| f(\mb{x}) - \hat{f}(\mb{x})\|^2}{\sum_{\mb{x} \in \mc{T}}{\|f(\mb{x})\|^2}}},
\]
where $\hat{f}(\x)$ denotes the approximation of $f(\x)$ by the proposed algorithms. 

We compare the sensitivity, the specificity, and the relative test error of our algorithms with those that are obtained from the standard neural network without regularization, the group Lasso neural network \cite{murray2015auto,murray2019autosizing}, and the monomial-based dictionary matrix method with sparse constraints \cite{brunton2016discovering, schaeffer2018extracting}.
To find the regularization hyperparameters of group Lasso and adaptive group Lasso, we use the popular Bayesian information criterion (BIC).
The remaining hyperparameters are fixed throughout the simulations. 
The learning rate of Adam, gradient descent, and the proximal algorithm is $0.005$.
Finally, all weights and coefficients of the final results whose absolute values are smaller than $10^{-4}$ are set to $0$.

\subsection{Data Simulation} 
\label{sec:sim}

To simulate the input data, we first solve the Lorenz-96 system \cite{lorenz1996predictability} numerically using \texttt{lsoda} method \cite{petzold1983automatic} with the time step size $\Delta t = 0.01$, from $t= t_0=0$ to $t= T_{final}=80$ and obtain the numerical solution $\{\widetilde{\x}(t_i)\}_{i=1}^m$:
\begin{equation}
\dfrac{\d x_j}{\dt} = -x_{j-2}\, x_{j-1} + x_{j-1} \, x_{j+1} - x_j + F,\quad j=1,\cdots, d.
\label{eqn:lorenz96}
\end{equation}
Here $x_0=x_d$ and $x_{d+1}=x_1$. In our simulations, we choose $d = 40$, $F = 8$, $m=\dfrac{T_{final} -t_0}{\Delta t} =8000$, and $t_i = i\, \Delta t$, for $i=1,\ldots, m$. The initial conditions are $x_j(0) = 1$ for all $j \ne 20$ and $x_{20}(0) = 1.008$. Then we add Gaussian noise with standard deviation $\sigma_x$ to the numerical solution of the Lorenz 96 and obtain the input data $\{\x_i\}_{i=1}^m$:
\[\x_i = \widetilde{\x}(t_i) + \varepsilon_xM_x,\]
where $M_x$ is the maximum value of the noiseless training inputs $t \in [0,8]$, and $\varepsilon_x\sim \mathcal{N}(0,\sigma_x^2 I_d)$. The simulated output data are
 \[y_i = f(\widetilde{\x}(t_i)) + \varepsilon_yM_y,\]
where $M_y$ is the maximum value of the noiseless training outputs $t \in [0,8]$ and $\varepsilon_y\sim \mathcal{N}(0,\sigma_y^2)$. The nonlinear function $f$ will be specified in each experiment. To validate those methods, we examine the learned functions on the validation set of 2000 samples $\{\x^{test}_i, y^{test}_i\}$, where $\{\x^{test}_i\}_{i=1}^{2000}$ is the numerical solution of the Lorenz 96 system from t = 80 to t = 100 and $y^{test}_i = f(\x^{test}_i)$.

\subsection{Effect of the Size of Neural Networks}

Here, we compare the performance of our proposed method with respect to $H$, the number of nodes in each hidden layer.
In this comparison, we assume that both inputs and outputs of the training data are noisy with $\sigma_x = \sigma_y = 0.02$.
We vary the number of nodes $H$ ($H =10,20,40,80$) in each hidden layer while keeping the size of the training set to be $8000$ as mentioned in Section \ref{sec:sim}.
For each $H$, we generate 100 different data sets with the same noise level and the same following target function:
\begin{equation*}
  f(\x) = -x_{23}x_{24} + x_{24}x_{26} - x_{25}+8,\end{equation*}
which is the right hand side of the $25^{th}$ equation of the Lorenz 96 system. The results are presented in Table \ref{Table1}.
In all cases, the sensitivities are one, which means every neural network is able to select all active variables. Also, the specificities are all greater than 0.8, which means all neural networks successfully remove most of the inactive variables. Moreover, the relative test errors are almost the same of around $6\%$, while the neural network with $H = 20$ provides the highest specificity of 92.7\%. Therefore, in the remaining of the paper, we fix the number of nodes in each hidden layer $H$ to be $20$.
\begin{table}[t!]
\begin{center}
\begin{tabular}{|c|c|c|c|c| }\hline
$H$ & \# parameters & Sensitivity & Specificity &  Rel. test error\\ \hline
10 & 681 & 1 & 0.807 & 0.051 \\
20 & 1681 &1 &0.927 & 0.062\\
40 & 4961&1 & 0.87 & 0.067\\ 
80 & 16321 &1 & 0.79 & 0.079\\ \hline
 \end{tabular}
\caption{The average sensitivities, specificities, and relative test errors of Algorithm \ref{alg} with different neural networks in approximating the function $f(\x) = -x_{23}x_{24} + x_{24}x_{26} - x_{25}+8$.}
\label{Table1}
\end{center}
\end{table}

\subsection{Effect of Noise Levels}

In this section, we test the performance of our proposed algorithm, the standard neural network without regularization, a group Lasso neural network method, and the sparse monomial-based dictionary method in approximating the following function
\[f(\x)= -x_8x_9 + x_9x_{11} -x_{10}+8,\] 
with various noise levels. This function is the governing equation of the 10-th equation of the Lorenz 96. 

\begin{figure}[htbp!]
\centering
\includegraphics[width = 3 in]{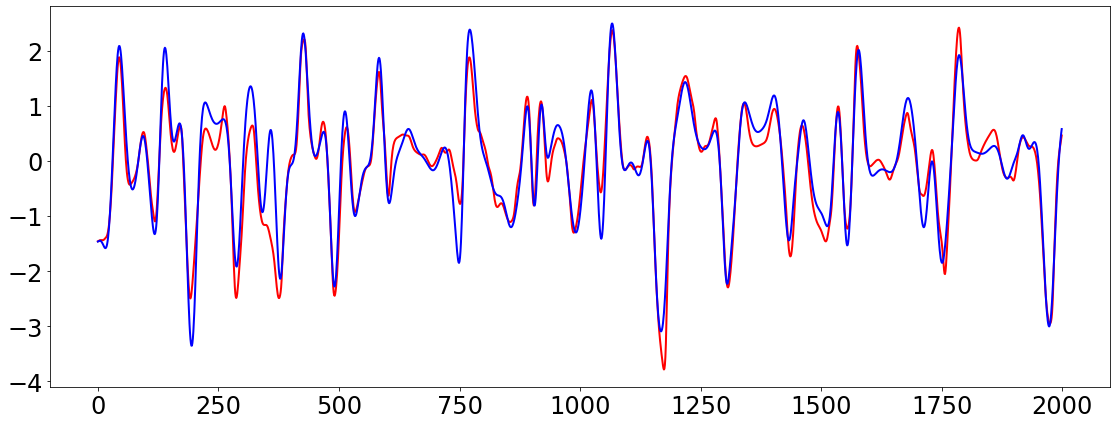}\quad
\includegraphics[width = 3 in]{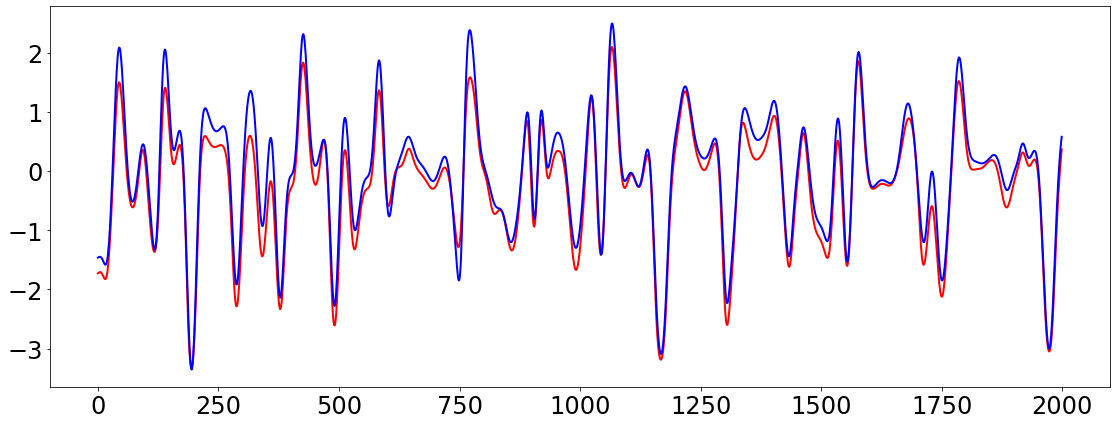}\\
\includegraphics[width = 3 in]{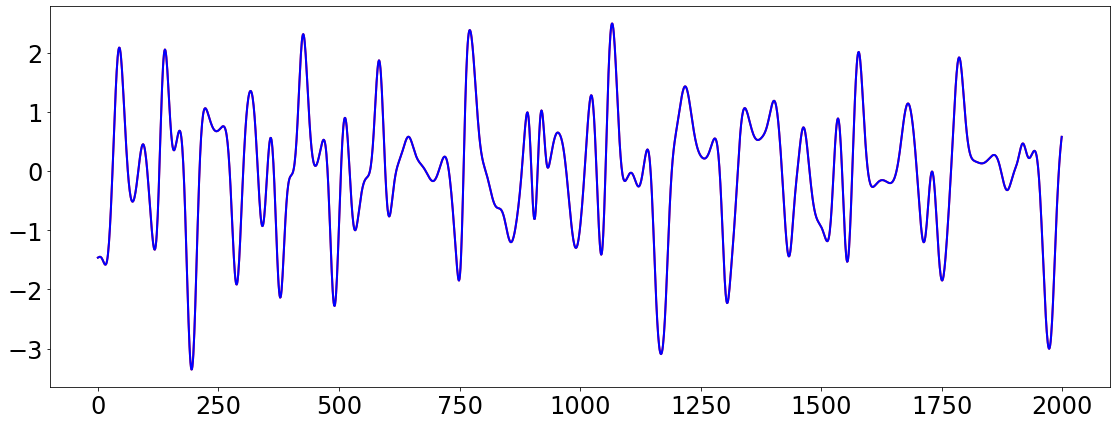}\quad
\includegraphics[width = 3 in]{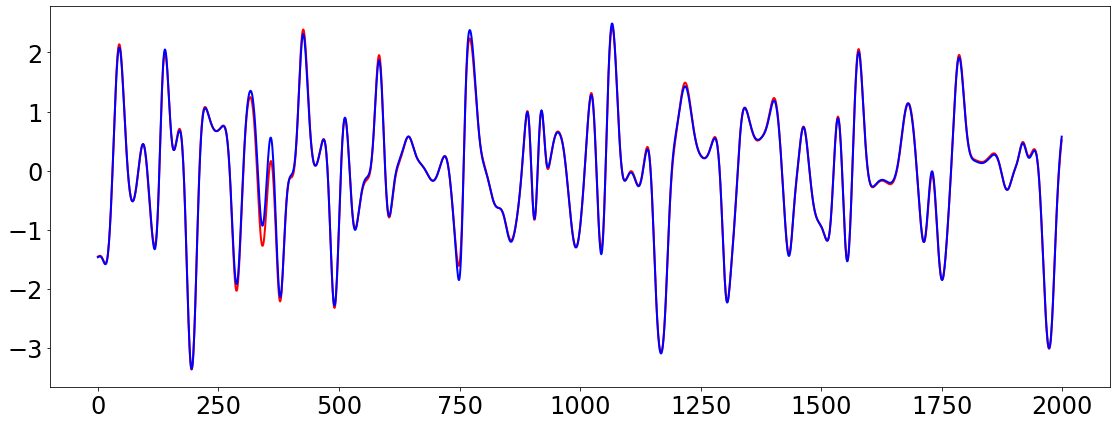}
\caption{Learned curves (in {\color{red} red}) of the standard neural network (top left), the group Lasso neural network (top right), the sparse monomial-based dictionary method (bottom left), and our method (bottom right) versus the ground truth curve (in {\color{blue}blue}) when there is no noise in both input and output data.}
\label{fig:Noisseless}
\end{figure}
\begin{figure}[htbp!]
\centering
\includegraphics[width = 3 in]{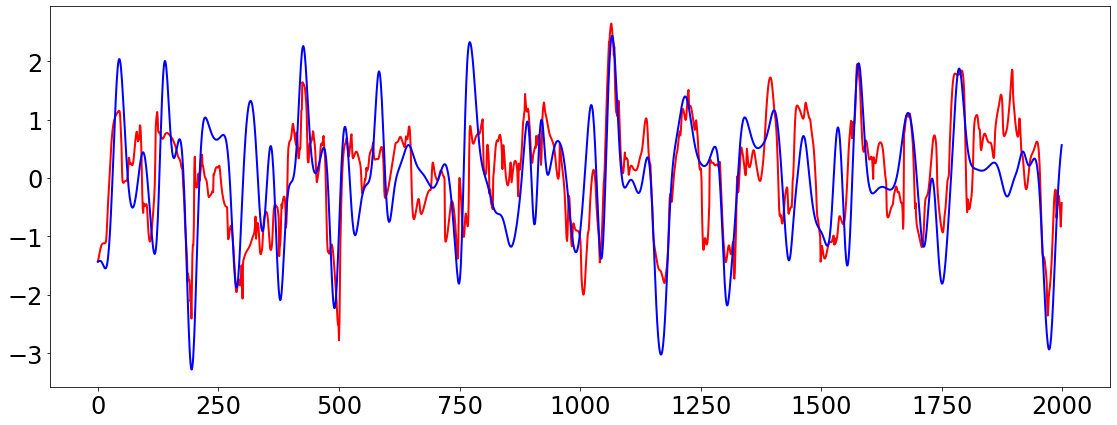}\quad
\includegraphics[width = 3 in]{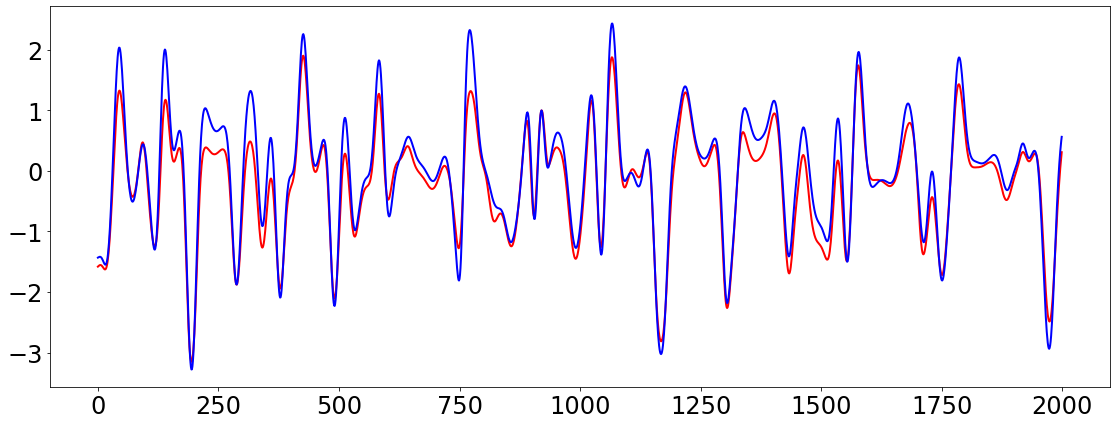}\\
\includegraphics[width = 3 in]{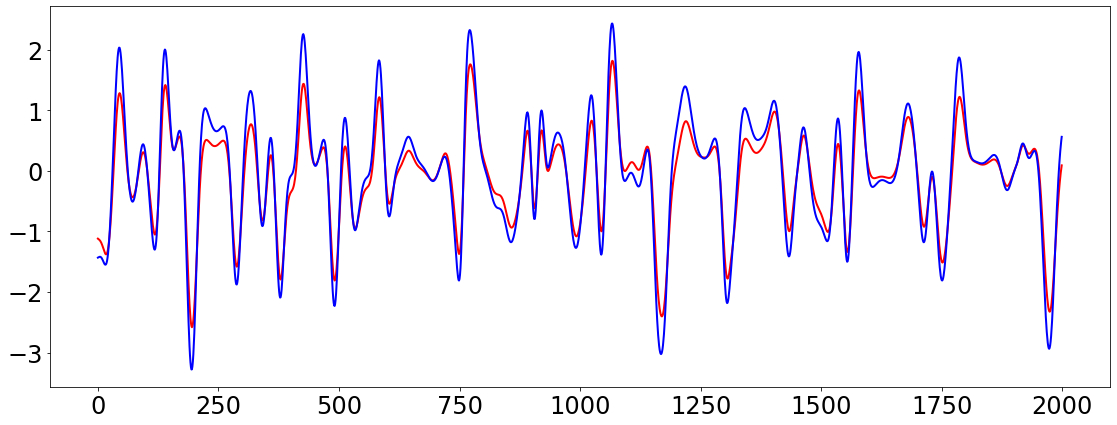}\quad
\includegraphics[width = 3 in]{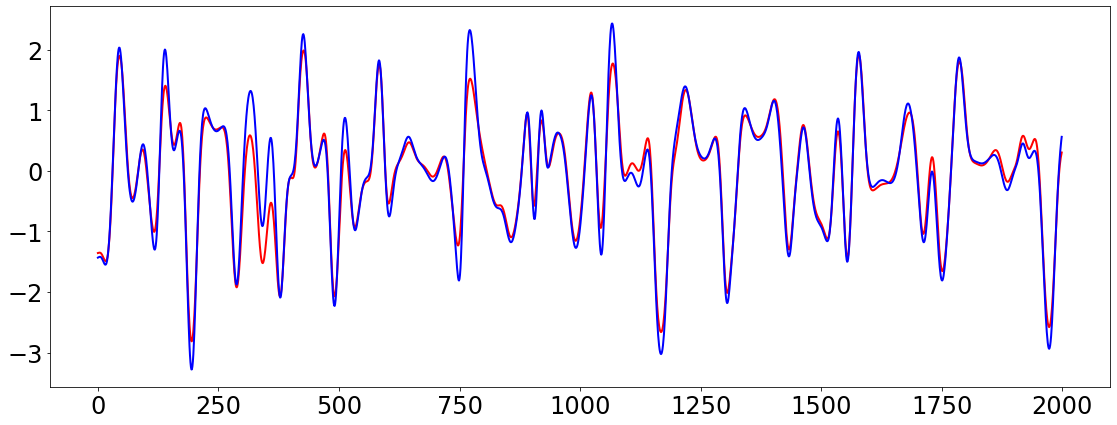}
\caption{Learned curves (in {\color{red} red}) of the standard neural network (top left), the group Lasso neural network (top right), the sparse monomial-based dictionary method (bottom left), and our method (bottom right) versus the ground truth curve (in {\color{blue}blue}) when the noise levels in input and output are $\sigma_x = \sigma_y = 0.04$.}
\label{fig:Noise004}
\end{figure}

When there is no noise in both input and output data, the monomial-based dictionary method and our method provide a better approximation than the other two methods. Specifically, the relative test errors are $0.33, 0.28,0.061,$ and $4.2\times 10^{-5}$ for the standard neural network without regularization, a group Lasso neural network method, the sparse monomial-based dictionary method, and our method, respectively. The learned curves are plotted in Figure \ref{fig:Noisseless}. In this scenario, the monomial-based dictionary method performs the best due to the fact that there is no noise and the true underlying function is polynomial. On the other hand, when both input and output data are noisy with noise level $\sigma_x = \sigma_y = 0.04$, our method performs slightly better than the monomial-based dictionary method.
The learned curves are shown in Figure \ref{fig:Noise004}. In this case, our method outperforms the other three methods with the smallest relative test error of $0.22$ and the sparse monomial-based dictionary method achieves the second smallest relative test error of $0.29$. 

\begin{figure}[htbp!]
\centering
\includegraphics[height = 2 in]{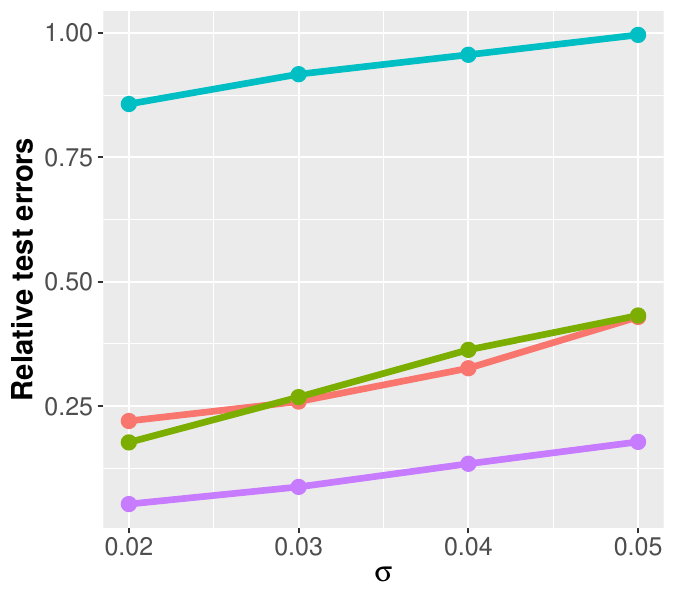}\label{fig:test_errors}\quad
\includegraphics[height = 2 in]{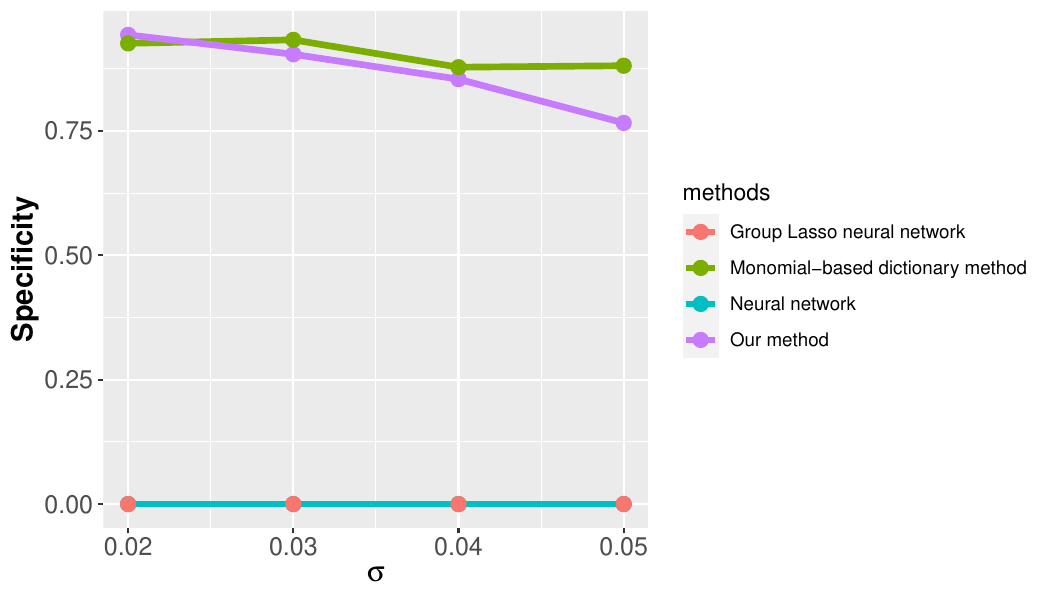}\label{fig:specificity}
\caption{Average relative test errors and average specificities of the standard neural network, the group Lasso neural network, the sparse monomial-based dictionary method, and our method across different noise levels, $\sigma_x = \sigma_y = \sigma = 0.02,0.03,0.04$, and $0.05$.}
\label{fig:results}
\end{figure}

We also perform a thorough simulation to study the effect of noise levels on these methods.
In this simulation, we vary the noise levels $\sigma_x = \sigma_y = 0.02, 0.03, 0.04, 0.05$ and for each noise level, we generate $100$ different data sets.
The average relative test errors for all methods are presented in Figure \ref{fig:results}.
We can see that our proposed method outperforms other methods for all noise levels. The relative test errors of the group Lasso neural network and sparse monomial-based dictionary method are similar. The standard neural network without regularization is the worse method, which emphasizes the importance of regularization in high dimensional problems. Moreover, the sensitivities of neural network and group Lasso neural network are always $1$ while their specificities are always $0$. That is, these models always include all variables in their model which aligns with the finding in \cite{NEURIPS2020_1959eb9d, ho2020consistent}.
On the other hand, the specificities of our model and the sparse monomial-based dictionary method are similar (see Figure \ref{fig:results}).
However, the sensitivity of our model is always $1$ while the sensitivity of the sparse monomial-based dictionary method decreases from $1$ to $0.8$ as the noise levels increase.
It means our model always captures all active variables and when the noise levels are high, the sparse monomial-based dictionary method often misses about $1$ out of $4$ active variables.

\subsection{Non-polynomial Functions}
\begin{figure}[t!]
\centering
\includegraphics[width = 3 in]{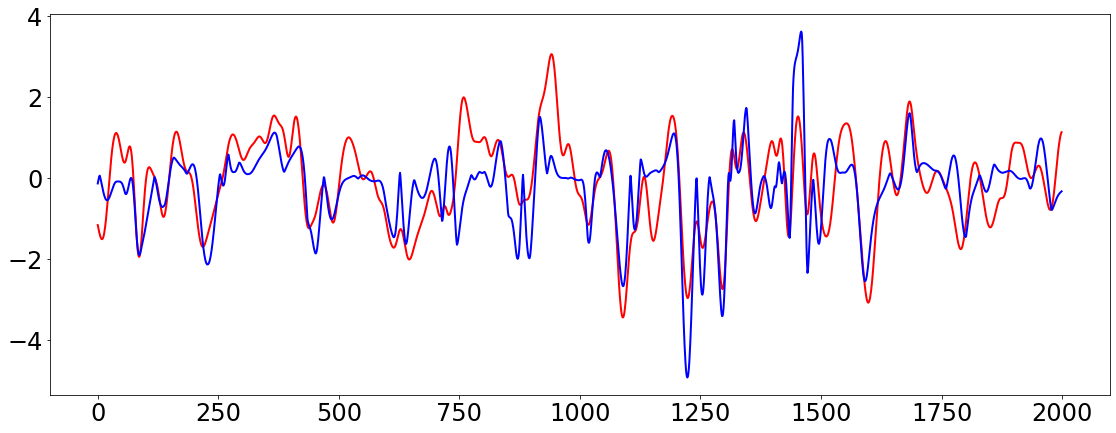}\quad
\includegraphics[width = 3 in]{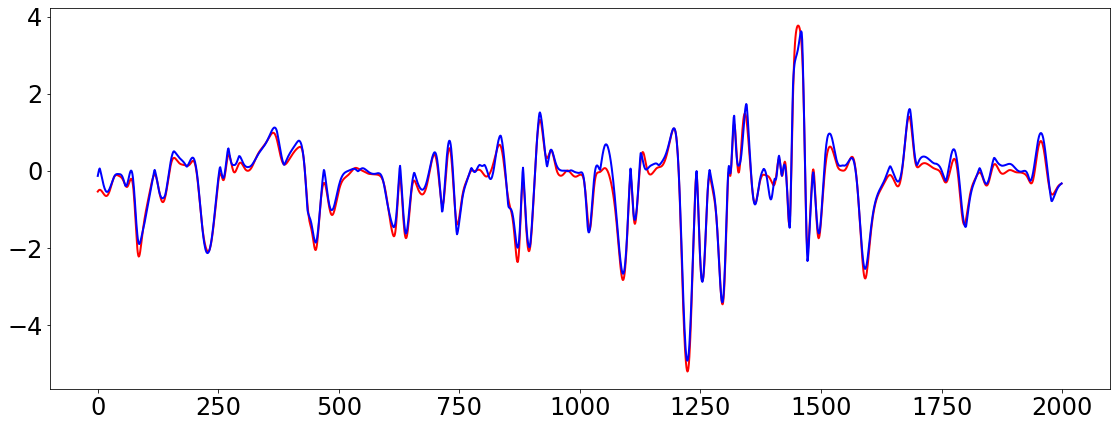}\\
\includegraphics[width = 3 in]{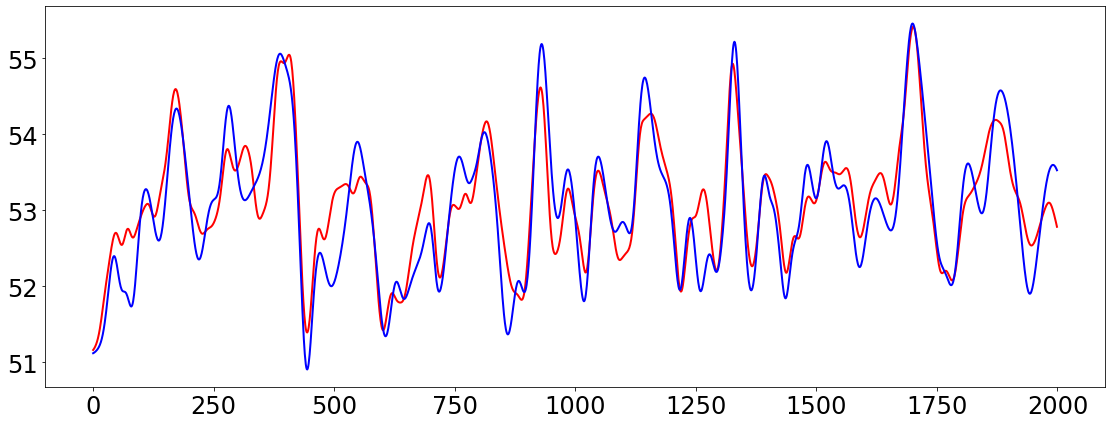}\quad
\includegraphics[width = 3 in]{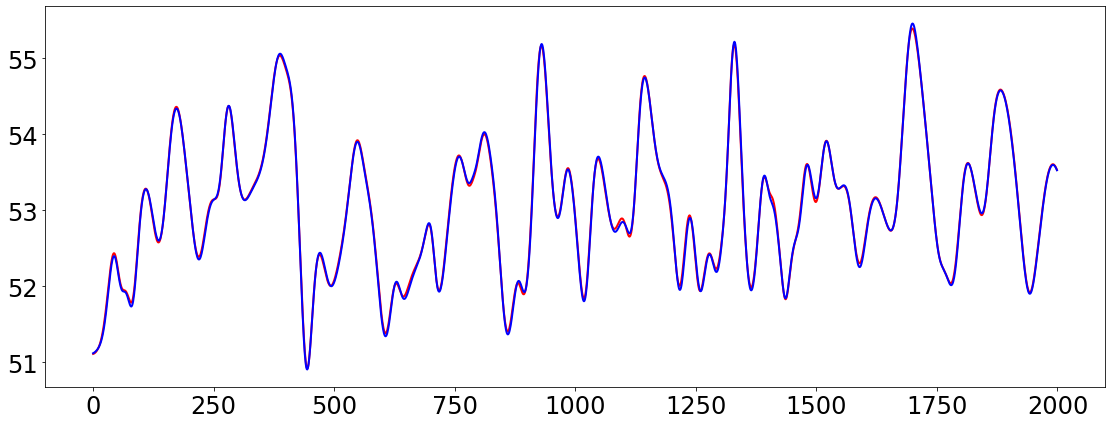}\\
\includegraphics[width = 3 in]{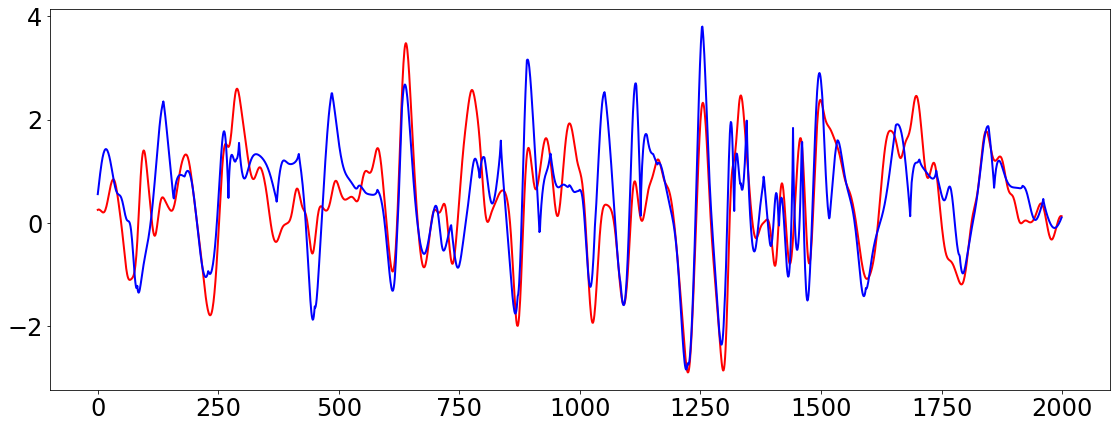}\quad
\includegraphics[width = 3 in]{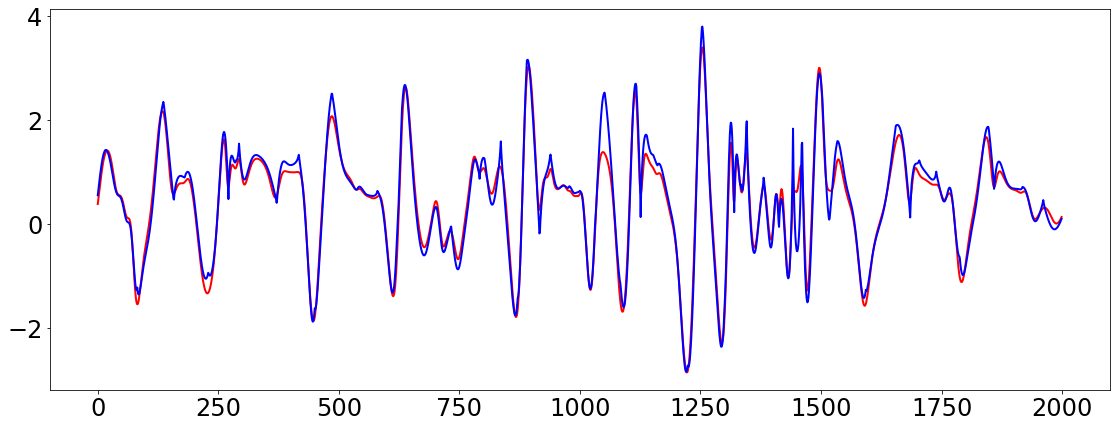}
\caption{Learned curves (in {\color{red}red}) of the sparse monomial-based dictionary method (left column) and of our method (right column) versus ground-truth curves (in {\color{blue}blue}) in Setting 1 (first row), Setting 2 (second row), and Setting 3 (third row).}
\label{fig:Nonlinear}
\end{figure}
Next, we examine the effectiveness of our algorithm to approximate non-polynomial functions.
Specifically, we consider the following settings.
\paragraph{Setting 1:}
The underlying function is
\[
f_1(\x)= (x_{19}^{4/3} - x_{16}^{4/3})x_{17}^{4/3} - x_{18}^{4/3} + 8.
\]
In this setting, the inputs are noiseless ($\sigma_x = 0$) while the outputs has Gaussian noise with standard deviation $\sigma_y = 0.02$.

\paragraph{Setting 2:}
The underlying function is
\[
f_2(\x)= (e^{x_{19}/50} - e^{x_{16}/50})e^{x_{17}/50} - e^{x_{18}/50} + 8.
\]
In this case, the outputs are noiseless ($\sigma_y = 0$) while the inputs has Gaussian noise with standard deviation $\sigma_x = 0.02$.

\paragraph{Setting 3:}
The underlying function is
\[
f_3(\x)= (e^{x_{19}/10} - x_{16}^{2/3})x_{17} - x_{18}^{4/5} + 8. 
\]
Here, both inputs and outputs are corrupted by noises ($\sigma_x = \sigma_y = 0.02$).

In all settings, $\x \in \mbb{R}^{40}$ is generated from the Lorenz-96 system as mentioned in Section \ref{sec:sim}. The learned curves are presented in Figure \ref{fig:Nonlinear} and the average relative rest errors over $100$ simulated data for each setting are shown in Table \ref{tab:test-nonpoly}.
In all settings, our method outperforms the sparse monomial-based dictionary method.
We note that relative errors of the second setting are smaller than those of other cases because the value of the underlying function is larger.
Additionally, the sensitivities and specificities of the sparse monomial-based dictionary method are always $1$ and $0$, respectively, indicating that this method cannot capture the active variables.
On the other hand, the sensitivities of our method are $1, 0.915, 1$ for three settings and the specificities are $0.889, 0.9997, 0.927$, respectively.

\begin{table}[t!]
\begin{center}
 \begin{tabular}{|c|c|c| } \hline
Setting &Our method & Sparse mononial-based dictionary method \\ \hline
1 & 0.141 & 0.844 \\ 
2 & 0.0009 & 0.007 \\ 
3 & 0.146& 0.622 \\ \hline
 \end{tabular}
\caption{Relative test errors of the sparse monomial-based dictionart method and of our method in approximating functions in Settings 1, 2, and 3.}
\label{tab:test-nonpoly}
\end{center}
\end{table}

\subsection{The case $f( \mb x) = g(A \mb x)$}

\begin{figure}[htbp!]
\centering
\includegraphics[width = 3 in]{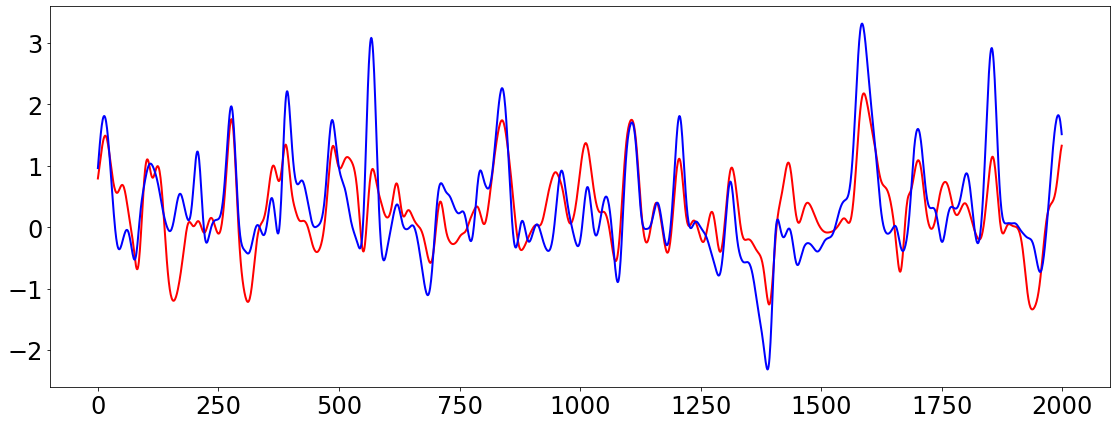}\quad
\includegraphics[width = 3 in]{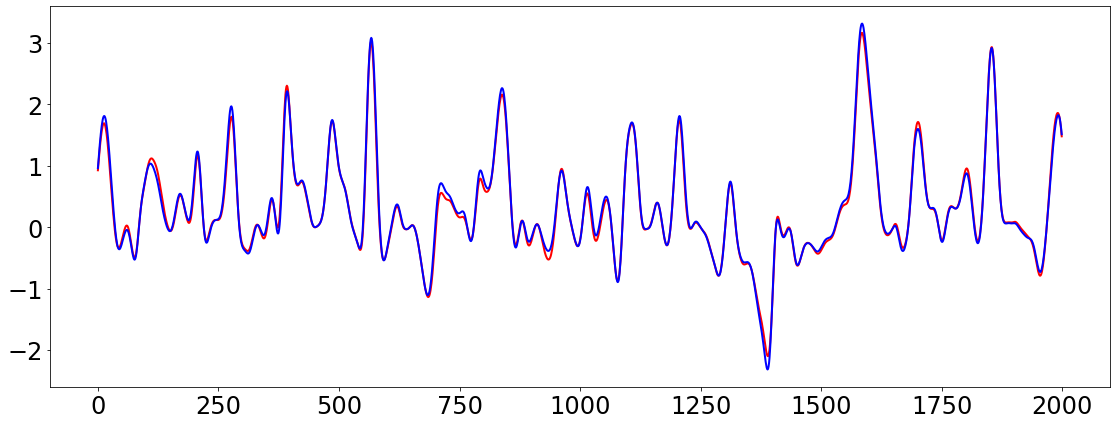}
\caption{Learned curve (in {\color{red}red}) of the sparse monomial-based dictionary method (left) and of our method (right) versus ground-truth curve (in {\color{blue}blue}) in approximating $f(\x) = g(A\x)$, where $g(\mathbf{z})= (z_4 - z_1)z_2 - z_3 + 8, \mathbf{z}=(z_1,z_2,z_3,z_4)$, and $A\in \R^{4\times 40}$.}
\label{fig:ModelReduction}
\end{figure}

In this experiment, we examine the underlying function $f(\x) = g(A \mb x)$ depends only on a few linear combinations of variables.
Here, the matrix $A \in \mbb{R}^{4 \times 50}$ is a Gaussian random matrix and $g: \mbb{R}^4 \to \mbb{R}$ is given as follows:
\[
g(\mathbf{z})= (z_4 - z_1)z_2 - z_3 + 8, \quad \mathbf{z} \in \mbb{R}^4.
\]
The inputs $\x \in \mbb{R}^{40}$ are generated from the Lorenz-96 system with no noise and the outputs are
\[
y = f(\x) + \varepsilon_y M_y = g(A \mb x) + \varepsilon_y M_y, \quad \varepsilon_y \sim \mc{N}(0,0.02^2).
\]
For this experiment, the training process is the same as previous sections.
The learned curves from our method and the sparse monomial-based dictionary method are visualized in Figure \ref{fig:ModelReduction}.
Clearly, our proposed method outperforms the sparse monomial-based dictionary method in approximating the underlying function.

\section{Conclusion}
We proposed a new neural network framework for high-dimensional function approximation with time-dependent data where the underlying function depends only on few active variables. By utilizing an adaptive group Lasso penalty to the weights of the first hidden layer of the neural network, our model can identify the active variables with high accuracy. We also proposed an adaptation of this framework to the scenario where the underlying function depends only on few linear combinations of variables. In this case, we apply the group Lasso penalty to the weights of the second (instead of the first) hidden layer of the neural network. Through various experiments, we show that our method outperforms recent popular function approximation methods including the sparse monomial-based dictionary method, the standard neural network, and the group Lasso neural network. 
In addition, we proved that the proposed optimization procedure is guaranteed to achieve loss decay.

\section*{Acknowledgement}
LSTH was supported by startup funds from Dalhousie University, the Canada Research Chairs program, the NSERC Discovery Grant, and the NSERC Discovery Launch Supplement. GT was supported by the NSERC Discovery Grant and the NSERC Discovery Launch Supplement. We thank Hans De Sterck for his helpful feedback which led to significant improvements.

\bibliographystyle{alpha}
\bibliography{refs}

\newpage
\appendix
\section{Proof of Theorem \ref{th:loss decay}}
\label{sec:convergence proof0}
 Let $\displaystyle p^{k+1} = \frac{\overline{\theta}^{k+1}-\theta^{k+1}}{\tau_{k}} \in \partial J(\theta^{k+1})$. Then 
\begin{equation}\label{eqn:gk}
g^k = -\dfrac{\overline{\theta}^{k+1} - \theta^k}{\tau_k} =- p^{k+1} - \dfrac{\theta^{k+1} -\theta^k}{\tau_k} .
\end{equation}
Using similar argument as in Theorem 3.2 \cite{bungert2021bregman} and in Lemma 1 \cite{yun2020general}, we have:
\begin{align*}
    & \L(\theta^{k+1}) - \L(\theta^k) \\
    & \leq \inp{\grad{\L(\theta^k})}{\theta^{k+1}-\theta^k} + \frac{C}{2}\norm{\theta^{k+1}-\theta^k}^2\\ 
    & = \inp{g^k}{\theta^{k+1}-\theta^k} + \inp{\grad{\L(\theta^k}) - g^k}{\theta^{k+1}-\theta^k} + \frac{C}{2}\norm{\theta^{k+1}-\theta^k}^2 \\     & =  \inp{p^{k+1}}{\theta^k - \theta^{k+1}} + \inp{\grad{\L(\theta^k}) - g^k}{\theta^{k+1}-\theta^k} + \left(\frac{C}{2} -\frac{1}{\tau_k}\right)\norm{\theta^{k+1}-\theta^k}^2\\
    & = - D_J^{p^{k+1}}(\theta^k,\theta^{k+1}) + J(\theta^k) - J(\theta^{k+1}) + \inp{\grad{\L(\theta^k}) - g^k}{\theta^{k+1}-\theta^k} + \left(\frac{C}{2} -\frac{1}{\tau_k}\right)\norm{\theta^{k+1}-\theta^k}^2,
\end{align*}
where the first inequality is due to the Lipschitz continuity of the function $\mathcal{L}$, the second equality comes from the substitution of $g^k$ (Equation \eqref{eqn:gk}), and the last equality is obtained from the Bregman distance formula:
\[D_J^{p^{k+1}}(\theta^k,\theta^{k+1}) = J(\theta^k) - J(\theta^{k+1}) - \inp{p^{k+1}}{\theta^k-\theta^{k+1}},\quad\text{for}\quad  p^{k+1}\in \partial J(\theta^{k+1}).\] 
Using Cauchy?Schwarz inequality, we have
\begin{align*}
    \mathcal{F}(\theta^{k+1})  +  D_J^{p^{k+1}}(\theta^k,\theta^{k+1}) + \left(\frac{1}{\tau_k} - \frac{C}{2}\right)\norm{\theta^{k+1}-\theta^k}^2   
    &\leq \mathcal{F}(\theta^k)+  \inp{\grad{\L(\theta^k}) - g^k}{\theta^{k+1}-\theta^k}  \\
    &  \leq \mathcal{F}(\theta^k)+ \dfrac{\nu}{2} \norm{\grad{\L(\theta^k}) - g^k}^2+ \dfrac{1}{2\nu} \norm{\theta^{k+1}-\theta^k}^2,
\end{align*}
for any $\nu>0.$ Since the Bregman distance $D_J^{p^{k+1}}(\theta^k,\theta^{k+1})$ is always nonnegative, we have
 \begin{equation*}
    \mathcal{F}(\theta^{k+1})  +   \left(\frac{1}{\tau_k} - \frac{C}{2} -\dfrac{1}{2\nu}\right)\norm{\theta^{k+1}-\theta^k}^2  \leq  \mathcal{F}(\theta^k) +  \dfrac{\nu}{2} \norm{\grad{\L(\theta^k}) - g^k}^2.
 \end{equation*}
 Choosing $\nu = \dfrac{1}{C}$ yields
 \begin{equation}\label{eqn:thm1a}
  \mathcal{F}(\theta^{k+1})+   \left(\frac{1}{\tau_k} - C\right)\norm{\theta^{k+1}-\theta^k}^2 \leq \mathcal{F}(\theta^k)+ \dfrac{1}{2C} \norm{g^k - \grad{\L(\theta^k})}^2.
\end{equation}
Taking the expectation with respect to the minibatches on both sides of Equation \eqref{eqn:thm1a}, we obtain Equation \eqref{eqn:theorem1}.
 
This proves the expectation of the function $\mathcal{F}(\theta) = \L(\theta) + J(\theta)$ decreases with each iteration, up to the variance of our gradient approximation $\grad{\mathbf{L}(\theta;\omega)}.$
\section{Proof of Theorem \ref{thm:convergence}} \label{sec:convergence proof}
\begin{proof}
From Equation \eqref{eqn:gk}, we have
\[ -\dfrac{1}{\tau_k} (\theta^{k+1} - \theta^k) - g^k +\nabla \mathcal{L}(\theta^{k+1})=p^{k+1} + \nabla \mathcal{L}(\theta^{k+1}) \in \partial\mathcal{F}(\theta^{k+1}).\]
Therefore,
\begin{align*}
    \dist(0, \partial\mathcal{ F}(\theta^{k+1}))^2 & = \inf_{v \in \partial \mathcal{F}(\theta^{k+1})}\norm*{0 - v}_2^2 \\
    & \leq \norm*{\frac{1}{\tau_k} (\theta^{k+1} - \theta^k) + g^k - \grad\L(\theta^{k+1})}_2^2 \\
    & = \norm*{g^k - \grad\L(\theta^k) + \grad\L(\theta^k) - \grad \L(\theta^{k+1}) + \frac{1}{\tau_k}(\theta^{k+1}-\theta^k)}_2^2 \\
    & \leq 3\norm*{g^k - \grad \L(\theta^k)}_2^2 + 3\norm*{\grad\L(\theta^k) - \grad \L(\theta^{k+1})}_2^2 + 3\norm*{\frac{1}{\tau_k}(\theta^{k+1}-\theta^k)}_2^2 \\
    & \leq 3\norm*{g^k - \grad \L(\theta^k)}_2^2 + 3 C^2\norm*{\theta^k - \theta^{k+1}}_2^2 + 3{\tau_k^{-2}}\norm*{\theta^{k+1}-\theta^k}_2^2 \\
    & \leq 3\norm*{g^k - \grad \L(\theta^k)}_2^2 + 3 \left(C^2 + (\tau_k)^{-2}\right)\norm*{\theta^{k+1} - \theta^k}_2^2 \\
    & \leq 3\norm*{g^k - \grad \L(\theta^k)}_2^2 + 3 \frac{C^2 + (\tau_k)^{-2}}{(\tau_k)^{-1} - C}\left(\dfrac{1}{2C} \norm{g^k - \grad{\L(\theta^k})}^2 + \mathcal{F}(\theta^k) - \mathcal{F}(\theta^{k+1})\right),
\end{align*}
where the last inequality is obtained from Equation \eqref{eqn:thm1a}.

Let $B = \max\limits_{t\in [\gamma_1,\gamma_2] }{\dfrac{t^{-2}+ C^2  }{t^{-1} - C}} =\max\left\{\dfrac{ \gamma_1^{-2} + C^2}{\gamma_1^{-1} - C}, \dfrac{ \gamma_2^{-2} + C^2}{\gamma_2^{-1} - C}\right\}<\infty$.
Then
\begin{align*}
    \dist(0, \partial \mathcal{F}(\theta^{k+1}))^2
    & \leq 3\norm*{g^k - \grad \L(\theta^k)}_2^2 + 3 B\left(\dfrac{1}{2C} \norm{g^k - \grad{\L(\theta^k})}^2 + \mathcal{F}(\theta^k) - \mathcal{F}(\theta^{k+1})\right) \\
    & \leq \left(3+\frac{3B}{2C}\right)\norm*{g^k - \grad \L(\theta^k)}_2^2 + 3 B\left(\mathcal{F}(\theta^k) - \mathcal{F}(\theta^{k+1})\right).
\end{align*}
Taking expectation with respect to the minibatches, we have
\[\mathbb{E}[ \dist(0, \partial \mathcal{F}(\theta^{k+1}))^2]\leq \left(3+\frac{3B}{2C}\right)\mu + 3B\mathbb{E}[\mathcal{F}(\theta^k) -  \mathcal{F}(\theta^{k+1})]. \]
Therefore,
\begin{align*}
    \frac{1}{K}\sum_{k=0}^{K-1}\mathbb{E} \left[\dist(0, \partial F(\theta^{k+1}))^2\right]
     \leq \left(3+\frac{3B}{2C}\right)\mu + \frac{3B}{K} \mathbb{E}[\mathcal{F}(\theta^0) - \mathcal{F}(\theta^K)].
\end{align*}

\end{proof}

\section{Proof of Theorem \ref{thm3}}\label{sec:lipschitz proof}
First we recall the definition of a Lipschitz function and introduce some notations.
\begin{mydef}
A function $f:V\rightarrow W$ between normed vector spaces $(V,\norm{\cdot}_V)$ and $(W,\norm{\cdot}_W)$ is $L_f$-Lipschitz if we have
\begin{equation}
    \norm{f(a) - f(b)}_W \leq L_f \norm{a - b}_V \quad \forall a,b\in V,
\end{equation}
where $L_f$ is a finite, non-negative real number.
\end{mydef}

\begin{mydef}
Let $f:V\rightarrow \R^{n \times p}$ be a bounded matrix-valued function between vector spaces $(V,\norm{\cdot}_V)$ and $(\R^{n \times p},\norm{\cdot}_F)$. Define
\begin{equation}
    M_f = \sup_{v\in V}\norm{f(v)}_F,
\end{equation}
where $\|A\|_F$ is the Frobenius norm of a matrix $A$.
\end{mydef}

\begin{mydef}
Let $f:V\rightarrow\R^{n \times p}$, $g:V\rightarrow\R^{p \times q}$ be matrix-valued functions defined on a normed vector space $(V,\norm{\cdot}_V)$. Define the product $f \ftimes g: V \rightarrow \R^{n \times q}$ by
\begin{equation}
    (f \ftimes g)(v) = f(v)g(v), \quad \forall v\in V,
\end{equation}
where $f(v)g(v)$ is the matrix product of the matrices $f(v)$ and $g(v)$.
\end{mydef}
We can easily compute a Lipschitz constant of the matrix-valued function $f\star g$.
\begin{lemma} \label{lemma:l matrix product}
Let $f:V\rightarrow\R^{n \times p}$, $g:V\rightarrow\R^{p \times q}$ be bounded and $L_f,L_g$-Lipschitz functions, respectively. Then $f\ftimes g$ is $\left(L_fM_g+L_gM_f\right)$-Lipschitz. 
\end{lemma} \label{lemma:l product}
\begin{cor} \label{cor1}
Let $f_i:V\rightarrow\R^{m_i \times m_{i+1}}$ be bounded and  $L_{f_i}$-Lipschitz. Then $F = f_1 \ftimes ... \ftimes f_N$ is Lipschitz with constant $L_F = \sum\limits_{i=1}^N \left(L_{f_i} \prod\limits_{j \neq i} M_{f_j}\right)$.
\end{cor}
\begin{proof}
This follows from repeatedly using Lemma \ref{lemma:l matrix product} on the functions $f_i$ and $(f_{i+1} \ftimes ... \ftimes f_N)$, and noting $M_{f\ftimes g} \leq M_f M_g$ for submultiplicative norms.
\end{proof}

We now detail our notation convention for derivatives involving vectors and matrices. 
\begin{mydef}
Let $\mathbf{y}:\R^d \rightarrow \R^n$ be a vector-valued function with a vector input. The derivative with respect to its input is the $n \times d$ Jacobian matrix
\[\partder{\mathbf{y}}{\mathbf{x}} = \begin{bmatrix} \partder{y_1}{x_1} & \hdots & \partder{y_1}{x_d} \\ \vdots & \ddots & \vdots \\  \partder{y_n}{x_1} & \hdots & \partder{y_n}{x_d} \end{bmatrix},\]
where $\mathbf{y} = [y_1,\ldots, y_n]^T$ and $\mathbf{x} =[x_1,\ldots, x_d]^T$.
\end{mydef}
Note that, if $n=1$, the derivative $\partder{\mathbf{y}}{\mathbf{x}}$ is the transpose of the gradient \[\partder{y}{\mathbf{x}} = \begin{bmatrix} \partder{y}{x_1} & \hdots & \partder{y}{x_d} \end{bmatrix} = \left( \grad y \right)^\T(x).\]
On the other hand, if $\mathbf{y} = \mat{A}\mathbf{x}$, we have $\partder{\mathbf{y}}{\mathbf{x}} = \mat{A}.$

\begin{mydef}\label{def2}
Let $y:\R^{n \times p} \rightarrow \R$ be a scalar valued function with a matrix input $\mat{X}$. The derivative with respect to its $n \times p$ matrix input is the $p \times n$ matrix
\[\partder{y}{\mat{X}} = \begin{bmatrix} \partder{y}{x_{11}} & \hdots & \partder{y}{x_{n1}} \\ \vdots & \ddots & \vdots \\  \partder{y}{x_{1p}} & \hdots & \partder{y}{x_{np}} \end{bmatrix}.\]
\end{mydef}

It is sometimes convenient to rewrite the affine function $T(\mathbf{x};\w,\mathbf{b}) := \w \mathbf{x}+\mathbf{b}$, where $\x\in\R^d, \mathbf{b}\in\R^h, \w\in\R^{h\times d}$, in terms of a single matrix $\widecheck{\w} \in \R^{h \times (d+1)}$ as follows:
\begin{equation*}
  T(\mathbf{x};\w,\mathbf{b}) = \widecheck{\w} \ \mathbf{\check{x}} = \begin{bmatrix}   & \w & \vline & \mathbf{b} \\  \end{bmatrix} \begin{bmatrix}  \mathbf{x} \\  1 \end{bmatrix}.
\end{equation*}

Using the chain rule, the partial derivatives of the one-hidden layer neural network function with respect to its weights and biases are given below.
\begin{prop} Under the assumptions in Theorem \ref{thm3}, 
the derivative of 
\[\l(\mathbf{x},\mathbf{y};\theta)= \half\norm{F_\theta(\mathbf{x}) - \mathbf{y}}_2^2\] with respect to its inner and outer layer parameters is given by
\begin{align*}
    \partder{\l(\theta)}{\widecheck{\w}_1} & = \check{\mathbf{x}}\mathbf{r}^\T\mat{W}_2 \diag(\sigma'(\mathbf{z})) \\
    \partder{\l(\theta)}{\widecheck{\w}_2} & = \mathbf{\check{s}}\mathbf{r}^\T,
\end{align*}
where $\mathbf{r} = F_\theta(\mathbf{x}) - \mathbf{y}$, $\mathbf{z} = \w_1 \mathbf{x} + \mathbf{b}_1$, $\mathbf{s} = \sigma(\mathbf{z})$, $\check{\mathbf{x}}$ to denote the vector $\mathbf{x}$ with an appended $1$, and $\check{\mathbf{s}}$ to denote the vector $\mathbf{s}$ with an appended $1$.
\end{prop}
We are now ready to prove Theorem \ref{thm3}.
\begin{proof}[Proof of Theorem 3]
For a fixed data point $(\mathbf{x},\mathbf{y})$, define the matrix-valued functions on $\theta = [\widecheck{\w}_{1},\widecheck{\w}_2]$:
\begin{enumerate}
    \item $G_1: \theta \mapsto [\sigma( \widecheck{\w}_1\mathbf{\check{x}})^T,  1] \in\R^{1 \times (h+1)}$
    \item $G_2: \theta \mapsto \widecheck{\w}_2^T  \in\R^{(h+1) \times n}$
    \item $G_3: \theta \mapsto \w_2 \in\R^{n \times h}$
    \item $G_4: \theta \mapsto \diag(\sigma'(\widecheck{\w}_1\mathbf{\check{x}})) \in\R^{h \times h}$
\end{enumerate}
Then 
\[f_1: = \partder{\l(\theta)}{\widecheck{\w}_1} = \check{\mathbf{x}} \ftimes (G_1\ftimes G_2-\mathbf{y}^T)\ftimes G_3\ftimes G_4(\theta)\quad\text{and}\quad
f_2:= \partder{\l(\theta)}{\widecheck{\w}_2} = G_1^\T \ftimes( G_1 \ftimes G_2 -\mathbf{y}^T) (\theta).\]
We compute Lipschitz constants and upper bounds of $G_i$ (see Definition \ref{def2}) for $i=1,2,3,4$:
\[M_{G_1} \leq \sqrt{h+1}, \quad M_{G_2} \leq C_1\sqrt{n(h+1)},\quad M_{G_3} \leq C_1\sqrt{n(h+1)},\quad M_{G_4} \leq \sqrt{h+1},
\]
\[L_{G_1} = \norm{\check{\mathbf{x}}}_2,\quad L_{G_2} = L_{G_3} = 1,\quad L_{G_4} = \norm{\check{\mathbf{x}}}_2.  \]
Using corollary \ref{cor1}, we obtain:
\begin{align*}
    L_{f_1} & = 2C_1\norm{\check{\mathbf{x}}}_2(h+1)^{3/2}\sqrt{n}\left(C_1\sqrt{n}\norm{\check{\mathbf{x}}}_2 + 1\right)\\
    L_{f_2} & = (h+1) \left(2 C_1\sqrt{n} \norm{\check{\mathbf{x}}}_2 + 1\right)\leq 2(h+1) \left( C_1\sqrt{n} \norm{\check{\mathbf{x}}}_2 + 1\right).
\end{align*}
Choose $L_{f_2} = 2(h+1) \left( C_1\sqrt{n} \norm{\check{\mathbf{x}}}_2 + 1\right)$.
Therefore, a Lipschitz constant for $\nabla_\theta \ell$ is 
\[
\begin{aligned}
L_{\grad_\theta\l}&=\sqrt{(L_{f_1})^2 + (L_{f_2})^2}= 2(h+1) \left( C_1\sqrt{n} \norm{\check{\mathbf{x}}}_2 + 1\right) \sqrt{C_1^2n(h+1) \|\check{x}\|_2^2 +1}\\
&= \mathcal{O}(nC_1^2h^{3/2}\|\check{x}\|_2^2),
\end{aligned}
\]
which completes the proof. 
\end{proof}
\end{document}